\documentclass{article}
\usepackage{spconf,amsmath,graphicx}
\usepackage[caption=false]{subfig}
\usepackage{amssymb}
\usepackage{multirow}
\usepackage{booktabs}
\usepackage{threeparttable}
\usepackage{bm}
\usepackage{fancyhdr}

\title{GaitMM: Multi-Granularity Motion Sequence Learning for Gait Recognition}
\name{Lei Wang \qquad Bo Liu$^{*}$\thanks{*This work was supported by the National Nature Science Foundation of China (61972132). Bo Liu is the corresponding author.} \qquad Bincheng Wang \qquad Fuqiang Yu}
\address{School of Information Science and Technology, Hebei Agricultural University, China}

\begin{document}
%
\maketitle

\thispagestyle{fancy}
\fancyhead{}
\lhead{}
\lfoot{\copyright~2023 IEEE}
\cfoot{}
\rfoot{}

%
\begin{abstract}

Gait recognition aims to identify individual-specific walking patterns by observing the different periodic movements of each body part. However, most existing methods treat each part equally and fail to account for the data redundancy caused by the different step frequencies and sampling rates of gait sequences. In this study, we propose a multi-granularity motion representation network (GaitMM) for gait sequence learning.  In GaitMM, we design a combined full-body and fine-grained sequence learning module (FFSL) to explore part-independent spatio-temporal representations. Moreover, we utilize a frame-wise compression strategy, referred to as multi-scale motion aggregation (MSMA), to capture discriminative information in the gait sequence.  Experiments on two public datasets, CASIA-B and OUMVLP, show that our approach reaches state-of-the-art performances. 
\end{abstract}
\section{Introduction}
\label{sec:intro}


Gait recognition has emerged as a promising biometric technology that leverages human gait information for long-distance identification without the cooperation of subjects. This technology has shown great potential in many fields, including video surveillance, rail transit, and sports simulation. However, gait recognition performance is often affected by various factors in real-world scenarios, such as changing viewpoints \cite{yu2006framework}, occlusion \cite{rida2019robust}, and different wearing conditions \cite{yeoh2016clothing,yao2021collaborative}. Therefore, learning gait representations that are invariant to these factors is a major challenge for gait recognition.

Most gait recognition methods utilize (convolutional neural networks) CNNs to extract spatio-temporal information from gait sequences. They can be categorized as set-based or sequence-based, depending on whether they consider the temporal order of frames. Set-based methods treat a gait sequence as an unordered set, which can either be compressed into a single gait template \cite{shiraga2016geinet} or learn order-independent gait representations from silhouette sets \cite{chao2021gaitset,hou2020gait}. Although the ordering of inputs is not essential for gait assessment in these methods, they may ignore the temporal nature of the gait sequence, resulting in the loss of discriminative local motion information.


Sequence-based methods tend to explore individual gait patterns from multiple spatial and temporal scales \cite{lin2020gait,fan2020gaitpart,lin2021gait,huang20213d}. As the input sequences are usually aligned, a uniform horizontal division of intermediate layer features can improve recognition performance \cite{fan2020gaitpart,lin2021gait}. Another approach introduces a body part-level localization module to achieve a more adaptive local representation \cite{huang20213d}. However, localization errors caused by changes in wear conditions or movement amplitude may degrade recognition accuracy. Additionally, the redundancy of adjacent frames limits the recognition of spatio-temporal variation patterns. While some methods \cite{lin2021gait,lin2021gaitmask} have been proposed to aggregate local clips, they may lack adaptability to motion aggregation.


\begin{figure*}[t]
\centering
\includegraphics[width=158mm]{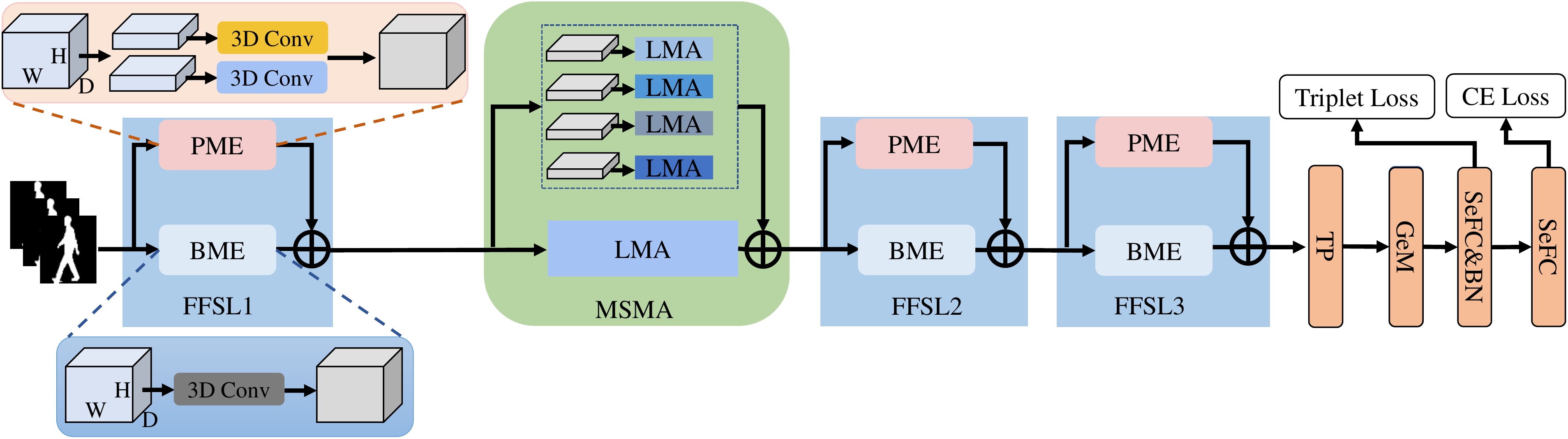}
\caption{Overview of GaitMM. The spatio-temporal dimensions of the feature map, i.e., $D$, $H$ and $W$, are indicated in the figure, and we omit the channel dimension $C$ for simplicity. The \bm{$\oplus$} represents the element-wise summation operation, and the TP represents the temporal pooling operation. The SeFC represents the separable fully connected layer.}
\label{fig:method}
\end{figure*}

To address the issues mentioned above, we propose a multi-scale motion learning framework named GaitMM for cross-view gait recognition. GaitMM comprises two main components: a combined full-body and fine-grained sequence learning module (FFSL) and a multi-scale motion aggregation (MSMA) operation. Rather than using shared convolutional kernels to extract part-specific features, in FFSL, the fine-grained motion patterns are independently obtained from body-part sequences. To reduce the redundancy of adjacent frames, MSMA compresses a sequence by aggregating information in each local clip.  The contributions of our work are summarised as follows:

    1) We propose a gait recognition framework named GaitMM, which combines global and fine-grained motion information for gait sequence learning. 
    
    
    2) We propose an adaptive MSMA module that reduces redundancy in the gait sequence.
    
    3) Experimental results on two public datasets, CASIA-B and OUMVLP, demonstrate that our method achieves state-of-the-art performance.

\section{Related Work}
\label{sec:format}
According to order sensitivity, there are two main categories of gait recognition techniques: set-based and sequence-based. In set-based approaches, gait silhouettes are typically considered as an unordered set, from which a set-level representation is obtained by characterizing the complementarity of the silhouettes in the set \cite{shiraga2016geinet,chao2021gaitset,hou2020gait,hou2021set,hou2022gait}. A straightforward way to handle a set of silhouettes is to compress them into a single template,  i.e., gait energy image (GEI), allowing the feature extraction and matching processes to be performed at the image level \cite{shiraga2016geinet}. However, these template-based methods largely ignore the spatial and temporal properties during preprocessing. In order to maximally preserve the set information, some methods take the raw silhouettes as inputs \cite{chao2021gaitset,hou2020gait,hou2022gait}. Chao et al. \cite {chao2021gaitset} first propose a set-based gait recognition framework named GaitSet, which employs a max-pooling function to learn a permutation-invariant representation of a set. Hou et al. \cite{hou2020gait} further propose a lateral connection to fuse silhouette-level and set-level features. While the methods above provide flexibility by dropping the sequential constraints, the temporal cues are also essential for revealing subtle gait changes.

Sequence-based approaches emphasize continuous pose variations, aggregating multi-scale motion features associated with body models or silhouettes. The model-based methods extract geometric and dynamic gait features from human motion models\cite{liao2020model,li2020end}. 
However, these approaches suffer from performance degradation caused by the inaccurate pose estimation results from low-resolution conditions. The silhouette-based methods usually extract spatio-temporal gait information from the sequence  \cite{lin2020gait,huang2021context,fan2020gaitpart,lin2021gait,huang20213d}.  To capture the various temporal cues in the sequence, some researchers considered extracting gait information from multiple temporal scales \cite{lin2020gait,huang2021context}. 
For example, 
Lin et al. \cite{lin2020gait} develop large and small temporal scales feature extractors for gait sequences using the designed 3D basic network blocks. %
Huang et al. \cite{huang2021context} explore the temporal features at three scales: frame-level, short-term and long-term. %
However, these methods insufficiently consider the motion differences among body parts. Therefore, some studies horizontally divide the silhouette into several parts and extract part-specific features \cite{fan2020gaitpart,lin2021gait,lin2021gaitmask}. %
 Moreover, Huang et al. \cite{huang20213d} propose 3D local operations to extract 3D volumes of body parts. Nevertheless, some irregular gait patterns  (such as wearing a coat) may affect the localization accuracy and reduce the recognition accuracy.

\section{Method}
\label{sec:pagestyle}


This section outlines the framework of our proposed method and describes several components, including the full-body and fine-grained sequence learning module (FFSL) and the multi-scale motion aggregation (MSMA) operation.

\subsection{Our Framework}



In GaitMM, multiple FFSL modules are stacked to learn gait motion features, and an MSMA module is available for frame-level downsampling. The whole pipeline is illustrated in Fig.~\ref{fig:method}. Given a gait sequence $\mathcal{S} \in \mathbb{R}^{C_{in}\times D \times H \times W}$, where $D$ means the number of frames, $(H,W)$ is the image size of each frame, $C$ denotes the number of  input channels. First, we feed $\mathcal{S}$ into GaitMM. Next, after frame compression by the MSMA module, the output feature map $\mathcal{F}_{FFSL3} \in \mathbb{R}^{C_{out}\times \frac{D}{3} \times H \times W}$ of the third FFSL modoule (FFSL3) is mapped to the discriminative space via temporal pooling (TP) \cite{lin2020gait,lin2021gait} and generalized mean pooling (GeM) \cite{lin2021gait,lin2021gaitmask} operations. Finally, we train the model using a combination of triplet and cross-entropy losses, which are commonly used for gait recognition \cite{lin2021gait,hou2020gait,huang20213d,chai2022lagrange}.

\subsection{Full-body and Fine-grained Sequence Learning}

\begin{table*}[tb]
    \begin{center}
    \caption{
Rank-1 accuracy (\%) on CASIA-B under all views and different conditions, excluding identical-view cases.
}
\label{table:casia-b}
    \setlength{\tabcolsep}{1.5mm}{
    \small
\begin{tabular}{c|c|c|ccccccccccc|c}
    \toprule
    \multicolumn{3}{c|}{Gallery NM \#1-4} & \multicolumn{11}{c|}{$0^{\circ}-180^{\circ}$} &\multirow{1}[4]{*}{Mean} \\
    \cline{1-14}
    \multicolumn{3}{c|}{Probe} & \multicolumn{1}{c}{$0^{\circ}$} & \multicolumn{1}{c}{$18^{\circ}$} & \multicolumn{1}{c}{$36^{\circ}$} & \multicolumn{1}{c}{$54^{\circ}$} & \multicolumn{1}{c}{$72^{\circ}$} & \multicolumn{1}{c}{$90^{\circ}$} & \multicolumn{1}{c}{$108^{\circ}$} & \multicolumn{1}{c}{$126^{\circ}$} & \multicolumn{1}{c}{$144^{\circ}$} & \multicolumn{1}{c}{$162^{\circ}$} & \multicolumn{1}{c|}{$180^{\circ}$}  \\
    \hline
    \multirow{13}[12]{*}{LT} & \multirow{5}[3]{*}{\shortstack{NM\\ \#5-6}} & GaitSet\cite{chao2021gaitset} & 90.8  & 97.9  & 99.4  & 96.9  & 93.6  & 91.7  & 95.0  & 97.8  & 98.9  & 96.8  & 85.8  & 95.0  \\ 
         &      
& GaitPart\cite{fan2020gaitpart} & 94.1  & 98.6  & 99.3  & 98.5  & 94.0  & 92.3  & 95.9  & 98.4  & 99.2  & 97.8  & 90.4  & 96.2  \\
         &       & GaitGL\cite{lin2021gait} & 96.0  & 98.3  & 99.0  & 97.9  & 96.9  & 95.4  & 97.0  & 98.9  & \textbf{99.3}  & 98.8  & 94.0  & 97.4  \\
                  &       & 3D Local\cite{huang20213d} & 96.0  & \textbf{99.0}  & \textbf{99.5}  & \textbf{98.9}  & \textbf{97.1}  & 94.2  & 96.3  & 99.0  & 98.8  & 98.5  & 95.2  & 97.5  \\
         &       & LagrangeGait\cite{chai2022lagrange} & 95.7  & 98.1  & 99.1  & 98.3  & 96.4  & 95.2  & 97.5  & 99.0  & \textbf{99.3}  & 98.9  & 94.9  & 97.5  \\
          &       & \textbf{Ours} & \textbf{97.2}  & 98.6  & 99.2  & 98.1  & 97.0  & \textbf{95.7}  & \textbf{97.8}  & \textbf{99.1}  & \textbf{99.3}  & \textbf{99.3}  & \textbf{96.6}  & \textbf{98.0}  \\
\cline{2-15} & \multirow{5}[3]{*}{\shortstack{BG\\ \#1-2}} & GaitSet\cite{chao2021gaitset} & 83.8  & 91.2  & 91.8  & 88.8  & 83.3  & 81.0  & 84.1  & 90.0  & 92.2  & 94.4  & 79.0  & 87.2  \\
        &       & GaitPart\cite{fan2020gaitpart} & 89.1  & 94.8  & 96.7  & 95.1  & 88.3  & 84.9  & 89.0  & 93.5  & 96.1  & 93.8  & 85.8  & 91.5  \\
         &       & GaitGL\cite{lin2021gait} & 92.6  & 96.6  & 96.8  & 95.5  & 93.5  & 89.3  & 92.2  & 96.5  & 98.2  & 96.9  & 91.5  & 94.5  \\
                   &       & 3D Local\cite{huang20213d} & 92.9  & 95.9  & \textbf{97.8}  & \textbf{96.2}  & 93.0  & 87.8  & 92.7  & 96.3  & 97.9  & \textbf{98.0}  & 88.5  & 94.3  \\
          &       & LagrangeGait\cite{chai2022lagrange} & 94.2  & 96.2  & 96.8  & 95.8  & 94.3  & 89.5  & 91.7  & 96.8  & 98.0  & 97.0  & 90.9  & 94.6  \\
         &       & \textbf{Ours} & \textbf{94.9}  & \textbf{97.1}  & 97.6  & 96.1  & \textbf{94.6}  & \textbf{91.2}  & \textbf{93.6}  & \textbf{97.4}  & \textbf{98.3}  & 97.0  & \textbf{93.3}  & \textbf{95.6}  \\
\cline{2-15}
        & \multirow{5}[3]{*}{\shortstack{CL\\ \#1-2}} & GaitSet\cite{chao2021gaitset} & 61.4  & 75.4  & 80.7  & 77.3  & 72.1  & 70.1  & 71.5  & 73.5  & 73.5  & 68.4  & 50.0  & 70.4  \\
          &       & GaitPart\cite{fan2020gaitpart} & 70.7  & 85.5  & 86.9  & 83.3  & 77.1  & 72.5  & 76.9  & 82.2  & 83.8  & 80.2  & 66.5  & 78.7  \\
         &       & GaitGL\cite{lin2021gait} & 76.6  & 90.0  & 90.3  & 87.1  & 84.5  & 79.0  & 84.1  & 87.0  & 87.3  & 84.4  & 69.5  & 83.6  \\
                  &       & 3D Local\cite{huang20213d} & 78.2  & 90.2  & 92.0  & 87.1  & 83.0  & 76.8  & 83.1  & 86.6  & 86.8  & 84.1  & 70.9  & 83.7  \\
         &       & LagrangeGait\cite{chai2022lagrange} & 77.4  & 90.6  & 93.2  & 90.2  & 84.7  & 80.3  & 85.2  & 87.7  & 89.3  & 86.6  & 71.0  & 85.1  \\
          &       & \textbf{Ours} & \textbf{81.3}  & \textbf{91.4}  & \textbf{93.7}  & \textbf{90.7}  & \textbf{86.8}  & \textbf{83.3}  & \textbf{86.2}  & \textbf{89.0}  & \textbf{91.8}  & \textbf{87.8}  & \textbf{76.9}  & \textbf{87.2}  \\
    \bottomrule
    \end{tabular}%
}
\end{center}
\end{table*}
The proposed FFSL module consists of a body-level motion feature extractor (BME) and a part-level motion feature extractor (PME). Specifically, the BME is implemented through a 3D convolution. Meanwhile, the PME learns part-independent spatio-temporal representations using non-shared 3D convolution filters that can account for diverse movement patterns of different body parts. For a input gait sequence $\mathcal{S}$, the process of BME can be formulated as:

\begin{equation}
    \mathcal{F}_{BME}=\mathrm{3DConv}_{3\times 3 \times 3}\left(\mathcal{S}\right),
\end{equation}
where $\mathrm{3DConv}_{3\times 3 \times 3}\left(\cdot\right)$ denotes a 3D convolution with a convolution kernel size of $3\times 3 \times 3$, $\mathcal{F}_{BME}\in \mathbb{R}^{C_{out}\times D \times {H_{}} \times W}$ is the output of the BME. For PME, the input $\mathcal{S}$ is evenly divided into $k$ parts along the horizontal axis, which are denoted as $\mathcal{S}^j,j\in 1,2,3,\cdots,k$, where $\mathcal{S}^j\in \mathbb{R}^{C_{in}\times D \times \frac{H_{}}{k} \times W}$. 
The PME process for $j\text{-th}$  part sequence is written as:
\begin{equation}
    \mathcal{F}_{PME}^j=\mathrm{3DConv}_{3\times 3 \times 3}\left(\mathcal{S}^j\right),
\end{equation}
where $\mathcal{F}_{PME}^j\in \mathbb{R}^{C_{out}\times D \times \frac{H_{}}{k} \times W}$ is the output feature map. Note that each part sequence undergoes a separate 3D convolution, ensuring independence and diversity of the learned spatio-temporal representations. Next, these part-level feature maps are concatenated along the horizontal axis, which can be formulated as:
\begin{equation}
    \mathcal{F}_{PME}=\mathcal{F}_{PME}^1\copyright\mathcal{F}_{PME}^2\cdots\copyright\mathcal{F}_{PME}^k,
    \label{equ:k}
\end{equation}
where $\copyright$ represents the concatenation operation , $\mathcal{F}_{PME}\in \mathbb{R}^{C_{out}\times D \times H \times W}$ is the output of PME. The output of FFSL is obtained by fusing $\mathcal{F}_{BME}$ and $\mathcal{F}_{PME}$ with an element-wise summation, which can be expressed as:
\begin{equation}
    \mathcal{F}_{FFSL}=\mathcal{F}_{BME}+\mathcal{F}_{PME}.
\end{equation}



\subsection{Multi-scale Motion Aggregation}
\label{sect:sta}

MSMA is employed to reduce data redundancy and enhance the discriminability of motions. It consists of two parallel branches, as shown in Fig.~\ref{fig:method}. Each branch is based on a local motion aggregation (LMA) operation, which is designed to perform temporal-downsampling for each gait sequence.  The part branch uses  $l$ separate LMAs for body parts to preserve distinctive movement patterns, while the global branch employs a body-level LMA to compress temporal information. The LMA can be formulated as:
\begin{equation}
    \mathcal{F}_{LMA}=p_1\textrm{Max}_{3\times1\times1}\left(\mathcal{F}\right)+p_2\textrm{Avg}_{3\times1\times1}\left(\mathcal{F}\right),
    \label{equ:p}
\end{equation}
where $\textrm{Max}_{3\times1\times1}\left(\cdot\right)$ denotes max pooling operation with kernel size ($3\times 1 \times1$), $\textrm{Mean}_{t\times1\times1}\left(\cdot\right)$ denotes average pooling operation with kernel size ($3\times 1 \times1$). $\mathcal{F}\in \mathbb{R}^{C_{in}\times D \times H \times W}$ and $\mathcal{F}_{LMA}\in \mathbb{R}^{C_2\times \frac{D}{3} \times H \times W}$ are the input and output of LMA, respectively. The $p_1$ and $p_2$ are two learnable parameters. 

\section{Experiments}

\subsection{Datasets and Implementation Details}


\textbf{CASIA-B.} The widely-used CASIA-B dataset \cite{yu2006framework} includes gait data for 124 subjects, captured from 11 camera views at regular intervals. Each view includes six normal walking (NM) sequences, as well as two sequences each of walking with a bag (BG) and walking with a coat (CL), resulting in a total of ten sequences per subject. Experiments in this study follow the large-sample training (LT) protocol \cite{chao2021gaitset}, in which the first 74 subjects are used for training and the remaining 50 for testing. During testing, NM\#01-04 sequences are used as the gallery, and NM\#05-06, BG\#01-02, and CL\#01-02 sequences are used as the probe for evaluation.


\noindent\textbf{OUMVLP.} The OUMVLP dataset \cite {NorikoTakemura2018MultiviewLP} is a large gait dataset, consisting of 10307 subjects. Each subject is captured at 14 camera views with a sampling interval of $15^{\circ}$, and each view includes two groups of sequences. Following the protocol in \cite{chao2021gaitset}, 5153 subjects are used for training and the remaining 5154 subjects for testing. During the testing phase, the sequences (Seq\#01) are regarded as the gallery, while the sequences (Seq\#00) are treated as the probe for evaluation.
\begin{figure}[t]
\centering
\includegraphics[width=69mm]{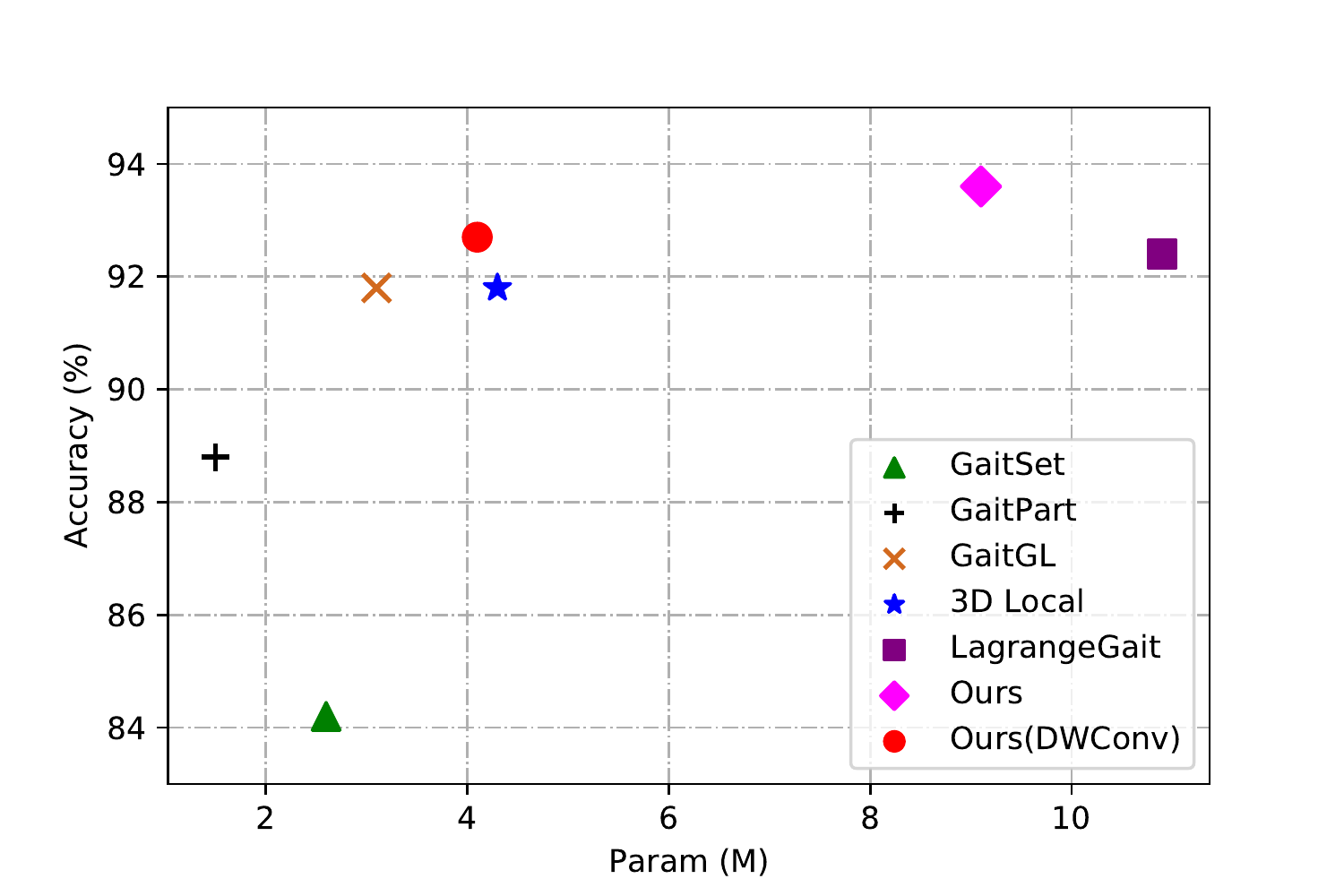}
\caption{The trade-off between accuracy and parameters of our method and other comparison methods on CASIA-B.}
\label{fig:trade}
\end{figure}

\noindent\textbf{Implementation Details.} The $k$ in Equ.~\ref{equ:k} and the $l$ in MSMA are both set to 8. The $p_1$ and $p_2$ in Equ.~\ref{equ:p} are initialized to 0.5.  The $\delta$ in GeM is initialized to 6.5, and the margin $\beta$ of the triplet loss is set to 0.2. The number of FFSL is set to 3 for CASIA-B and double for OUMVLP. The gait silhouettes are aligned as  \cite{NorikoTakemura2018MultiviewLP} and the silhouette images uniformly crop to a size of $64\times 44$. The batch size $\left(P \times K\right)$ is set to $\left(8,8\right)$ on CASIA-B and $\left(32,8\right)$ on OUMVLP. During training, the number of frames $D$ is set to 30, and the model uses a Adam optimizer with the initial learning rate of 1$e$-4. For CASIA-B, the number of iterations is 80K, and the learning rate reset to 1e-5 after 70K. For OUMVLP, the number of iterations is 160K, the learning rate reset to 1$e$-5 after 150K iterations. 
\subsection{Comparison with State-of-the-Art Methods}
\textbf{CASIA-B.}
Tab.~\ref{table:casia-b} shows the performance comparison of our proposed GaitMM with the state-of-the-art (SOTA) methods on CASIA-B. Our approach achieves mean view recognition accuracies of 98.0\%, 95.6\% and 87.2\% for the NM, BG and CL walking conditions, respectively, which are  0.5\%, 1.0\% and 2.1\% higher than LagrangeGait \cite{chai2022lagrange}, demonstrating the superiority of GaitMM in cross-view recognition.  However, the operation of independent feature extraction in FFSL increases the number of parameters.  To address this issue, we replace the 3D convolution in PME with the 3D depthwise separable convolution (DWConv) \cite{chollet2017xception}. As shown in Figure \ref{fig:trade}, we can observe that the proposed methods, especially the DWConv version, achieve a better trade-off between model size and accuracy.


\noindent\textbf{OUMVLP.}
Tab.~\ref{table:oumvlp} presents the rank-1 accuracy of GaitMM evaluated on OUMVLP compared to several SOTA methods. The results demonstrate that our proposed method outperforms the current methods in all views, highlighting the generalization capability of GaitMM.

\subsection{Ablation Study} 
The effects of FFSL and MSMA are shown in Table \ref{tab:abla}. Removing both PME and MSMA leads to a decrease in performance. FFSL is necessary  for accurately modeling spatial scale information and capturing motion relationships between body parts, while MSMA is important for extracting discriminative temporal clues while compressing gait sequences.

\begin{table}[htbp]
  \centering
  \caption{Rank-1 accuracy (\%) on OUMVLP under all views, excluding identical-view cases.}
  \label{table:oumvlp}
  \setlength{\tabcolsep}{0.6mm}{
  \small
    \begin{tabular}{c|cccc|c}
    \toprule
    \multirow{2}[4]{*}{Probe} & \multicolumn{5}{c}{Gallery All 14 views} \\
\cline{2-6}          & GaitSet\cite{chao2021gaitset} & GaitPart\cite{fan2020gaitpart} & GaitGL\cite{lin2021gait}& 3D Local\cite{huang20213d} & \textbf{Ours} \\
    \hline
    0°    & 84.5  & 88.0  & 90.5  & -  & \textbf{92.9} \\
    
    15°   & 93.3  & 94.7  & 96.1  & -  & \textbf{97.1} \\
   
    30°   & 96.7    & 97.7  & 98.0  & -  & \textbf{98.4} \\
    
    45°   & 96.6  & 97.7    & 98.1  & -  & \textbf{98.4} \\
   
    60°   & 93.5    & 95.5  & 97.0  & -  & \textbf{97.5} \\
    75°   & 95.3  & 96.6  & 97.6  & -    & \textbf{98.0} \\
    90°   & 94.2  & 96.2  & 97.1  & -  & \textbf{97.7} \\
    180°  & 87.0  & 90.6  & 94.2  & -  & \textbf{95.8} \\
    195°  & 92.5  & 94.2  & 94.9  & -  & \textbf{96.3} \\
    210°  & 96.0    & 97.2    & 97.4  & -  & \textbf{97.8} \\
    225°  & 96.0  & 97.1  & 97.4  & -  & \textbf{97.8} \\
    240°  & 93.0  & 95.1    & 95.7  & -  & \textbf{96.4} \\
    255°  & 94.3  & 96.0  & 96.5  & -  & \textbf{97.1} \\
    270°  & 92.7  & 95.0  & 95.7  & -  & \textbf{96.6} \\
    \hline
    Mean  & 93.3  & 95.1  & 96.2  & 96.5  & \textbf{97.0} \\
    \bottomrule
    \end{tabular}%
    }
\end{table}%

\begin{table}[htbp]
  \centering
  \caption{Ablation study on FFSL  and MSMA.}
  \small
    \begin{tabular}{c|c|c|c|c|c|c}
    \toprule
    \multicolumn{2}{c|}{FFSL} & \multirow{2}[1]{*}{MSMA} & \multicolumn{4}{c}{Rank-1 Accuracy} \\
\cline{1-2}\cline{4-7}    BME   & PME   &       & NM    & BG    & CL    & Mean \\
    \hline
    \checkmark     &       &       & 97.1  & 94.4  & 84.1  & 91.9  \\
    \checkmark     & \checkmark     &       & 97.8  & 95.2  & 85.2  & 92.7  \\
    \checkmark     &       & \checkmark     & 97.1  & 94.0  & 85.1  & 92.1  \\
    \checkmark     & \checkmark     & \checkmark     & \textbf{98.0}  & \textbf{95.6}  & \textbf{87.2}  & \textbf{93.6}  \\
    \bottomrule
    \end{tabular}%
  \label{tab:abla}
\end{table}%


\section{Conclusion}

This paper proposes GaitMM, a novel gait recognition framework that integrates fine-grained and global motion properties. The FFSL module is designed to learn the part-based sequence and body representations, while the MSMA operation aggregates sequence information by compressing redundant frames. We conduct extensive experiments on two public datasets to demonstrate the effectiveness of GaitMM.

\clearpage

\bibliographystyle{IEEEbib}
\bibliography{Template}

\begin{thebibliography}{10}

\bibitem{yu2006framework}
Shiqi Yu, Daoliang Tan, and Tieniu Tan,
\newblock ``A framework for evaluating the effect of view angle, clothing and
  carrying condition on gait recognition,''
\newblock in {\em 18th International Conference on Pattern Recognition (ICPR)},
  2006, vol.~4, pp. 441--444.

\bibitem{rida2019robust}
Imad Rida, Noor Almaadeed, and Somaya Almaadeed,
\newblock ``Robust gait recognition: a comprehensive survey,''
\newblock {\em IET Biometrics}, vol. 8, no. 1, pp. 14--28, 2019.

\bibitem{yeoh2016clothing}
TzeWei Yeoh, Hern{\'a}n~E Aguirre, and Kiyoshi Tanaka,
\newblock ``Clothing-invariant gait recognition using convolutional neural
  network,''
\newblock in {\em 2016 International Symposium on Intelligent Signal Processing
  and Communication Systems (ISPACS)}, 2016, pp. 1--5.

\bibitem{yao2021collaborative}
Lingxiang Yao, Worapan Kusakunniran, Qiang Wu, Jingsong Xu, and Jian Zhang,
\newblock ``Collaborative feature learning for gait recognition under cloth
  changes,''
\newblock {\em IEEE Transactions on Circuits and Systems for Video Technology},
  vol. 32, pp. 3615--3629, 2021.

\bibitem{shiraga2016geinet}
Kohei Shiraga, Yasushi Makihara, Daigo Muramatsu, Tomio Echigo, and Yasushi
  Yagi,
\newblock ``Geinet: View-invariant gait recognition using a convolutional
  neural network,''
\newblock in {\em ICB}, 2016, pp. 1--8.

\bibitem{chao2021gaitset}
Hanqing Chao, Kun Wang, Yiwei He, Junping Zhang, and Jianfeng Feng,
\newblock ``Gaitset: Cross-view gait recognition through utilizing gait as a
  deep set,''
\newblock {\em IEEE Transactions on Pattern Analysis and Machine Intelligence},
  vol. 44, pp. 3467--3478, 2021.

\bibitem{hou2020gait}
Saihui Hou, Chunshui Cao, Xu~Liu, and Yongzhen Huang,
\newblock ``Gait lateral network: Learning discriminative and compact
  representations for gait recognition,''
\newblock in {\em European Conference on Computer Vision (ECCV)}, 2020, pp.
  382--398.

\bibitem{lin2020gait}
Beibei Lin, Shunli Zhang, and Feng Bao,
\newblock ``Gait recognition with multiple-temporal-scale 3d convolutional
  neural network,''
\newblock in {\em Proceedings of the 28th ACM International conference on
  Multimedia}, 2020, pp. 3054--3062.

\bibitem{fan2020gaitpart}
Chao Fan, Yunjie Peng, Chunshui Cao, Xu~Liu, Saihui Hou, Jiannan Chi, Yongzhen
  Huang, Qing Li, and Zhiqiang He,
\newblock ``Gaitpart: Temporal part-based model for gait recognition,''
\newblock in {\em Proceedings of the IEEE/CVF conference on computer vision and
  pattern recognition (CVPR)}, 2020, pp. 14225--14233.

\bibitem{lin2021gait}
Beibei Lin, Shunli Zhang, and Xin Yu,
\newblock ``Gait recognition via effective global-local feature representation
  and local temporal aggregation,''
\newblock in {\em Proceedings of the IEEE/CVF International Conference on
  Computer Vision (ICCV)}, 2021, pp. 14648--14656.

\bibitem{huang20213d}
Zhen Huang, Dixiu Xue, Xu~Shen, Xinmei Tian, Houqiang Li, Jianqiang Huang, and
  Xian-Sheng Hua,
\newblock ``3d local convolutional neural networks for gait recognition,''
\newblock in {\em Proceedings of the IEEE/CVF International Conference on
  Computer Vision (ICCV)}, 2021, pp. 14920--14929.

\bibitem{lin2021gaitmask}
Beibei Lin, Yu~Liu, and Shunli Zhang,
\newblock ``Gaitmask: Mask-based model for gait recognition,''
\newblock in {\em 32nd British Machine Vision Conference (BMVC)}, 2021, pp.
  1--12.

\bibitem{hou2021set}
Saihui Hou, Xu~Liu, Chunshui Cao, and Yongzhen Huang,
\newblock ``Set residual network for silhouette-based gait recognition,''
\newblock {\em IEEE Transactions on Biometrics, Behavior, and Identity
  Science}, vol. 3, no. 3, pp. 384--393, 2021.

\bibitem{hou2022gait}
Saihui Hou, Xu~Liu, Chunshui Cao, and Yongzhen Huang,
\newblock ``Gait quality aware network: Toward the interpretability of
  silhouette-based gait recognition,''
\newblock {\em IEEE Transactions on Neural Networks and Learning Systems},
  2022.

\bibitem{liao2020model}
Rijun Liao, Shiqi Yu, Weizhi An, and Yongzhen Huang,
\newblock ``A model-based gait recognition method with body pose and human
  prior knowledge,''
\newblock {\em Pattern Recognition}, vol. 98, pp. 107069, 2020.

\bibitem{li2020end}
Xiang Li, Yasushi Makihara, Chi Xu, Yasushi Yagi, Shiqi Yu, and Mingwu Ren,
\newblock ``End-to-end model-based gait recognition,''
\newblock in {\em Proceedings of the Asian conference on computer vision
  (ACCV)}, 2020, pp. 3--20.

\bibitem{huang2021context}
Xiaohu Huang, Duowang Zhu, Hao Wang, Xinggang Wang, Bo~Yang, Botao He, Wenyu
  Liu, and Bin Feng,
\newblock ``Context-sensitive temporal feature learning for gait recognition,''
\newblock in {\em Proceedings of the IEEE/CVF International Conference on
  Computer Vision (ICCV)}, 2021, pp. 12909--12918.

\bibitem{chai2022lagrange}
Tianrui Chai, Annan Li, Shaoxiong Zhang, Zilong Li, and Yunhong Wang,
\newblock ``Lagrange motion analysis and view embeddings for improved gait
  recognition,''
\newblock in {\em Proceedings of the IEEE/CVF Conference on Computer Vision and
  Pattern Recognition}, 2022, pp. 20249--20258.

\bibitem{NorikoTakemura2018MultiviewLP}
Noriko Takemura, Yasushi Makihara, Daigo Muramatsu, Tomio Echigo, and Yasushi
  Yagi,
\newblock ``Multi-view large population gait dataset and its performance
  evaluation for cross-view gait recognition,''
\newblock {\em Ipsj Transactions on Computer Vision and Applications}, vol. 10,
  pp. 1--14, 2018.

\bibitem{chollet2017xception}
Fran{\c{c}}ois Chollet,
\newblock ``Xception: Deep learning with depthwise separable convolutions,''
\newblock in {\em Proceedings of the IEEE conference on computer vision and
  pattern recognition}, 2017, pp. 1251--1258.

\end{thebibliography}

\end{document}